\documentclass{preprint}
\usepackage{pdfpages}

\usepackage[numbers,comma,sort&compress]{natbib}
\usepackage[utf8]{inputenc} 
\usepackage[T1]{fontenc}    
\usepackage{hyperref}       
\usepackage{url}            
\usepackage{booktabs}       
\usepackage{amsfonts}       
\usepackage{nicefrac}       
\usepackage{microtype}      
\usepackage{graphicx}

\usepackage{xcolor}         

\title{EvidenceOutcomes: a Dataset of Clinical Trial Publications with Clinically Meaningful Outcomes}

\author[1,2,$\dagger$]{Yiliang Zhou}
\author[3,$\dagger$]{Abigail M. Newbury}
\author[3]{Gongbo Zhang}
\author[3]{Betina Ross Idnay}
\author[4]{Hao Liu}
\author[3,*]{Chunhua Weng}
\author[1,*]{Yifan Peng}
\affil[1]{Weill Cornell Medicine}
\affil[2]{University of California, Irvine}
\affil[3]{Columbia University}
\affil[4]{Montclair State University}
\affil[*]{Corresponding author(s). Email(s): \url{yip4002@med.cornell.edu}, \url{cw2384@cumc.columbia.edu}}
\affil[$\dagger$]{These authors contributed equally to this work.}

\begin{document}

\maketitle

\begin{abstract}
  The fundamental process of evidence extraction and synthesis in evidence-based medicine involves extracting PICO (Population, Intervention, Comparison, and Outcome) elements from biomedical literature. However, Outcomes, being the most complex elements, are often neglected or oversimplified in existing benchmarks. To address this issue, we present EvidenceOutcomes, a novel, large, annotated corpus of clinically meaningful outcomes extracted from biomedical literature. We first developed a robust annotation guideline for extracting clinically meaningful outcomes from text through iteration and discussion with clinicians and Natural Language Processing experts. Then, three independent annotators annotated the Results and Conclusions sections of a randomly selected sample of 500 PubMed abstracts and 140 PubMed abstracts from the existing EBM-NLP corpus. This resulted in EvidenceOutcomes with high-quality annotations of an inter-rater agreement of 0.76. Additionally, our fine-tuned PubMedBERT model, applied to these 500 PubMed abstracts, achieved an F1-score of 0.69 at the entity level and 0.76 at the token level on the subset of 140 PubMed abstracts from the EBM-NLP corpus. EvidenceOutcomes can serve as a shared benchmark to develop and test future machine learning algorithms to extract clinically meaningful outcomes from biomedical abstracts. 
\end{abstract}

\section{Background \& Summary}

\begin{figure}
  \centering
  \includegraphics[width=.9\linewidth]{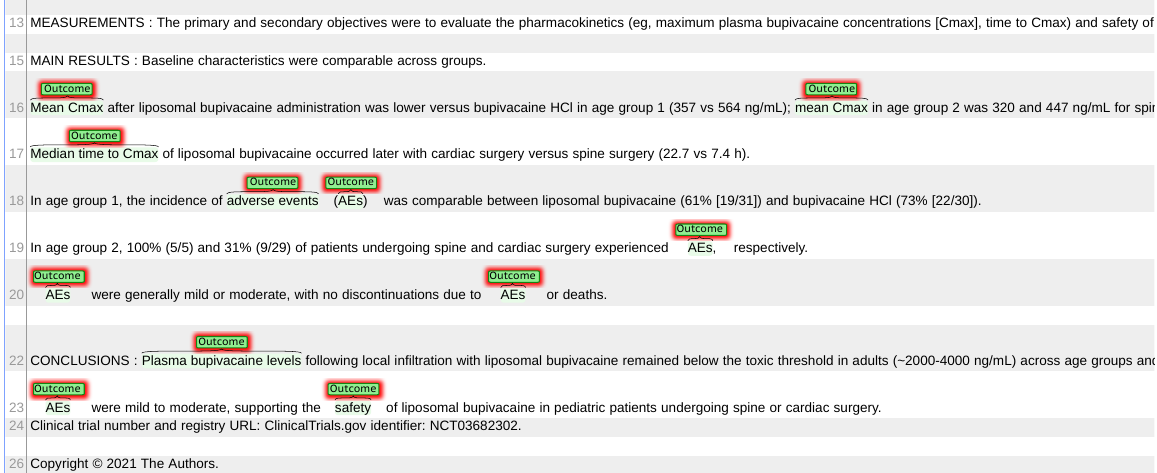} 
  \caption{Annotation examples from EvidenceOutcomes using BRAT.}
  \label{fig:example}
\end{figure}

The PICO (Population, Intervention, Comparison, and Outcome) framework has been widely adopted to help healthcare professionals structure clinical evidence extraction queries~\cite{Richardson1995-wi}. Figure~\ref{fig:example} illustrates examples from EvidenceOutcomes using BRAT to annotate the PICO elements. The extraction of PICO elements for downstream tasks is primarily performed by automated named entity recognition (NER) models in natural language processing (NLP)~\cite{Frandsen2020-mn, Brockmeier2019-gb, Wang2022-vz, kang2019pretraining, Zhang2024Span}. For efficient training of these NER models, high-quality PICO element annotations are crucial. However, three unresolved issues hinder the development of PICO extraction models.

First, the annotations for Outcome elements in existing corpora have often shown poor inter-annotator agreement compared to other elements~\cite{Nye2018-ul}. Sometimes, the annotations were even completely dropped due to overall low inter-annotator agreement~\cite{Zlabinger2018-tw}. Yet, the accurate extraction of Outcome elements is extremely valuable. These elements are crucial in updating clinical guidelines and, subsequently, clinical procedures based on Outcome element extraction from randomized controlled trial (RCT) literature~\cite{Uhlig2016-mo}. Furthermore, the extraction of Outcome elements helps organizations such as the Patient-Centered Outcomes Research Institute (PCORI) in assessing if the evaluated results cater to patients' needs, and Core Outcome Measures in Effectiveness Trials (COMET) in judging if the trials meet their specified core outcome set~\cite{Pcori2021-et, COMET_Initiative2023-jh}. Hence, annotating high-quality Outcome elements for training NLP models plays a significant role in improving the practice of evidence-based medicine.

Second, existing PICO annotation guidelines need to adequately consider the clinical meaningfulness of the Outcome elements. Some guidelines opt to annotate only those sentences that contain Outcomes rather than entity-level outcomes, justifying it with low inter-annotator agreement or the intrinsic difference of outcomes compared to other PICO elements\cite{demner2005knowledge, Boudin2010-nw}. For example, Demner-Fushman \textit{et al.}\cite{demner2005knowledge} annotated each sentence in an abstract whether it stated an outcome or not. Others occasionally neglect clinically pertinent attributes of outcomes and inadvertently capture unrelated outcome elements from abstract sections providing no evidence information, such as Background or Future work sections, thereby clouding the clarity of the results~\cite{Nye2018-ul}. For instance, the rules in evidence-based medicine (EBM)-NLP overlook words like `reduced' in Outcomes, resulting in the omission of key findings~\cite{Nye2018-ul}. As seen in the PubMed abstract with PMID 30953107, ``LDL-C reduction $\geq$50\%'' is an endpoint of the study~\cite{Rosenson2019-wo}.

Finally, it is noteworthy that while existing corpora have entity-level outcome element annotations, they usually cater to specific domains. For example, the corpus developed by Sanchez-Graillet \textit{et al.} is limited to abstracts pertaining to glaucoma and type 2 diabetes~\citet{Sanchez-Graillet2022-xi}. Similarly, the EBM-NLP corpus is exclusive to specific areas such as cardiovascular disease, cancer, and autism~\cite{Nye2018-ul}.

To address these challenges, we created EvidenceOutcomes, a carefully curated corpus of 500 randomly selected PubMed abstracts and an additional 140 abstracts drawn from the EBM-NLP. EvidenceOutcomes is an open-access corpus compiled to assist researchers in designing and developing NER models. To promote transparency and usability, we provide an extensive breakdown of the curation methods and benchmark developments pertinent to EvidenceOutcomes. We also identify its limitations and potential avenues for future exploration and welcome user feedback for further enhancement. 
%


\section{Methods}

\subsection{Ethics}

EvidenceOutcomes contains abstracts from two sources: PubMed and the EBM-NLP corpus. Since both sources are in the public domain, this study does not need institutional review board approval. The dataset is released under MIT licencse.

\subsection{Data collection strategy}

\begin{figure}
\centering
\includegraphics[width=.6\linewidth]{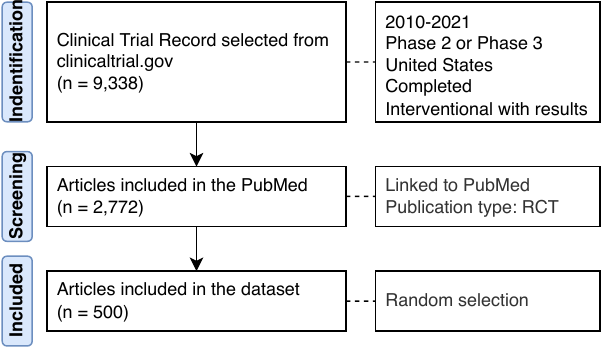}
\caption{Data collection procedure}
\label{fig:collection}
\end{figure}

A total number of 500 RCT abstracts were collected from PubMed and 140 abstracts from the EBM-NLP corpus. Figure~\ref{fig:collection} illustrates the data collection procedure for the 500 RCT abstracts. The selection criteria included (a) all clinical trials conducted from 2010 to 2021, (b) trials either in their Phase 2 or Phase 3, (c) completed interventional studies with available results, and (d) studies that were registered on \url{ClinicalTrials.gov} and conducted in the United States. This selection process resulted in 9,338 unique studies, each with a distinct National Clinical Trial (NCT) number. These studies were then linked to their respective articles on PubMed. Specifically, we used Entrez Programming Utilities to search for PubMed articles referencing the NCT number either in the title or abstract or Secondary Source ID, and with the term `Randomized Controlled Trial' as the Publication Type~\cite{Sayers2022-xm}. In cases where multiple PubMed articles were linked to a single NCT number, we selected the most recent one. This data collection strategy resulted in 3,066 NCT numbers with 2,772 unique PubMed articles (PMIDs). From them, a random selection of 500 PMIDs was made. During the annotation process, if an article was identified as a study protocol or trial design, it was replaced by another abstract selected at random.

An additional 140 abstracts from the EBM-NLP corpus were also randomly selected and the outcome elements were reannotated. This step was taken to enhance the annotation guidelines' comprehensiveness, improve the generalizability of the NLP models trained on EvidenceOutcomes, and allow for comparability with existing work. These specific abstracts were randomly selected from well-structured RCTs that were not study protocols or trial designs, had Results and Conclusions sections, and were associated with an NCT number.

\subsection{Annotation process}

\begin{figure}
\centering
\includegraphics[width=.9\linewidth]{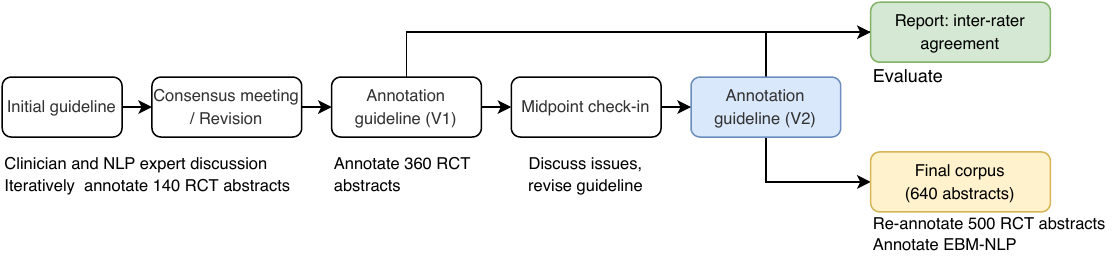}
\caption{Workflow of the annotation process}
\label{fig:workflow}
\end{figure}

We utilized an iterative annotation process (Figure~\ref{fig:workflow}). We first developed the annotation guideline based on sample abstracts from PubMed and existing definitions of clinical trial outcomes. Consultations with clinicians and NLP experts were utilized to integrate clinical and domain-specific expertise into the guideline.

Next, we refined the guideline through a trial-and-error process using batches of 10, 30, and 92 abstracts. For each abstract, the Outcome elements were identified by one primary annotator (AN) who worked on all abstracts, and two secondary annotators (BI and HL) who each worked on random non-overlapping subsets of abstracts. The annotation focused solely on the Results and Conclusions sections. After each batch, the annotators compared, discussed the results, and refined the guidelines further. Once the 132 abstracts were annotated, the guideline was finalized and dubbed "version 1". To ensure all annotators fully understood this version, an additional eight abstracts were annotated.

Subsequently, the remaining 360 abstracts were annotated in four separate batches using the established ``version 1''. After each batch, the annotators conferred to reach a consensus.

Upon finalizing the annotations, three key updates were made to the guideline, leading to ``version 2''. The first amendment created more flexibility for terms such as ``increase'' and ``improvement''. The second laid out a more precise protocol for annotating outcomes, mentioning ``baseline'' (Rule 8 in Supplementary File 1). Specifically, we only annotated `from baseline' if ``changes in,'' ``differences in,'' ``improvement in,'' or a similar entity is also included in the outcome (\textit{e.g.}, ``change in serum Cr level from baseline'' or ``differences in TAHC from baseline''). The third one clarified guidelines for annotating conjunction ``or'' as a mathematical symbol (\textit{e.g.}, ``grade 3 or worse''). These revisions addressed areas of confusion and disagreement commonly encountered in version 1. In accordance with these changes, 500 RCT abstracts were further revised, and 140 abstracts from the EBM-NLP corpus were newly annotated according to version 2.

\section{Data Records}

We used BRAT (Brat Rapid Annotation Tool, \url{https://brat.nlplab.org/}), which is a web-based annotation tool designed for collaborative text annotation tasks, particularly in the field of NLP and machine learning, to create the annotations.

The free text of the selected PubMed abstract, brat configuration files, and annotated data files are all available at \url{https://github.com/ebmlab/EvidenceOutcomes}.

\paragraph{Free-text (\texttt{.txt}) Files}

Extracted free-text abstracts from the 640 RCT publications. Each text file follows the naming format:\texttt{[PMID].txt} (PubMed identifier). The first line is the title of the article. The following lines contain the abstract of the article.

\paragraph{Annotation (\texttt{.ann}) Files}

Annotations in Brat's native ANN output format. Each annotation file follows the naming format: \texttt{[PMID].ann}. As per the ANN format, each line corresponds to a single entity. For each entity, the fields are as follows: item ID (\textit{e.g.}, T1), entity type (i.e., Outcome), string start index (\textit{e.g.}, 1852), string end index (\textit{e.g.}, 1858), and text (\textit{e.g.}, `stroke'). In order to visualize the annotations, we need to install Brat and unzip the entire zip file into the folder. Instructions for downloading and installing brat are available at \url{http://brat.nlplab.org/}.

\paragraph{Configuration (\texttt{.conf}) Files}

These are the brat configuration files (\texttt{annotation.conf}, \texttt{visual.conf}, \texttt{tools.conf}, and \texttt{kb\_shortcuts.conf}) used to control the annotation type, display, tool, and keyboard shortcut in a brat project. A detailed description of the brat annotation configuration is available at \url{https://brat.nlplab.org/configuration.html}.

\section{Technical Validation}

\subsection{Inter-annotator agreement}

The inter-annotator agreement was assessed on 360 abstracts from the 500 RCT corpus from PubMed with guideline version 1 and on 140 abstracts from the EBM-NLP corpus with guideline version 2 before reaching any consensus. Table \ref{tab:kappa} shows that the annotation process resulted in substantial agreement amongst annotators~\cite{McHugh2012-ha}.
\begin{table}
    \centering
    \caption{Cohen's Kappa score for annotation batches}
    \begin{tabular}{lrr}
    \toprule
    & 360 abstracts from PubMed & 140 abstracts from EBM-NLP\\
    \midrule
w/secondary annotator 1 & 0.733 & 0.760\\
w/secondary annotator 2 & 0.764 & 0.713\\
\hspace{1em}\textit{Average} & 0.748 & 0.738\\
    \bottomrule
    \end{tabular}
    \label{tab:kappa}
\end{table}

On the 360 abstracts from PubMed, Cohen's Kappa scores were 0.733 and 0.764 when comparing the annotations from the primary annotator with those from secondary annotators 1 and 2, respectively, resulting in an average of 0.748. As on the 140 abstracts from EBM-NLP, Cohen's Kappa scores were 0.760 and 0.713, respectively, yielding an average of 0.738.

Though the average Cohen's Kappa for guideline version 2 is slightly lower than that for version 1, significant updates incorporated in version 2 were crucial, as noted in the Methods section. The given Cohen's Kappa scores suggest a substantial level of shared understanding among the annotators regarding the definition of outcome elements as outlined in the annotation guideline~\cite{Craggs2005-bi}.

The annotation guidelines provide an approach for annotating Outcome entities in RCT abstracts (see Supplementary File 1). It includes 15 primary rules, each with specific elements to ensure consistency, clarity, and specificity through the annotation process. Broadly, the guidelines cover five main areas: when and how to annotate Outcome entities; the scope of outcomes to consider; particular details and formats that should be included or excluded while annotating Outcome entities; specific terms and phrases that should be annotated or omitted; and logistics and cross-reference procedures.

\subsection{Descriptive statistics}

\begin{figure}
\centering
\vspace{1em}
\includegraphics[width=\linewidth]{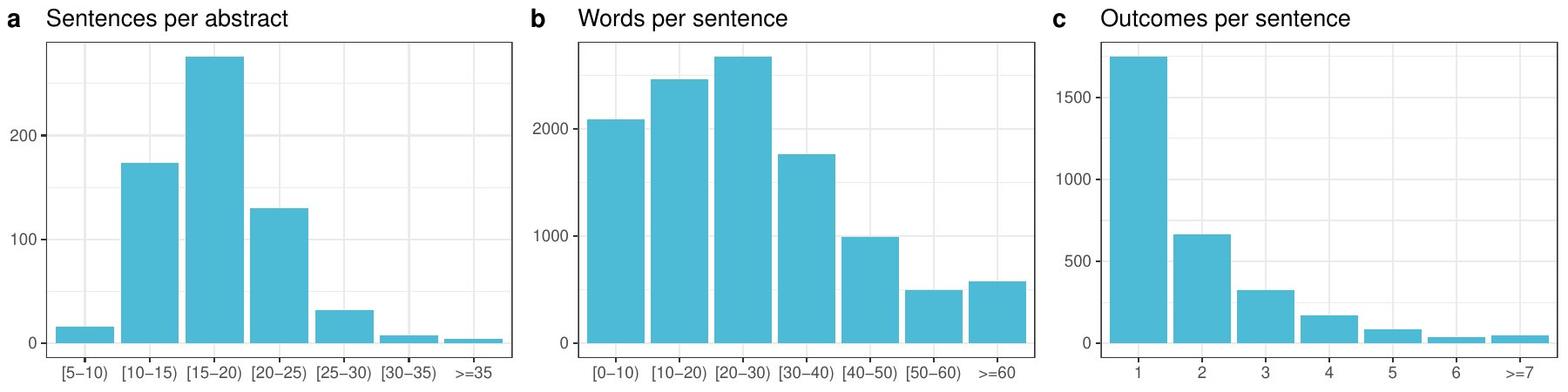}
\caption{Statistics of EvidenceOutcomes. (a) Sentences per abstract. (b) Words per sentence. (c) Outcomes per sentence}
\label{fig:Statistics}
\end{figure}

Figure~\ref{fig:Statistics} shows the distributions of sentences per abstract, words per sentence, and outcomes per sentence on EvidenceOutcomes. Notably, a substantial portion of the abstracts contain between 10 to 25 sentences (Figure~\ref{fig:Statistics}a). In contrast, a considerable number of sentences contain 10 to 40 words each (Figure~\ref{fig:Statistics}b). Figure~\ref{fig:Statistics}c illustrates the distribution of outcomes per sentence in our corpus. A significant portion of the sentences encompasses 1 Outcome each.

\begin{figure}
\centering
\vspace{1em}
\includegraphics[width=.8\linewidth]{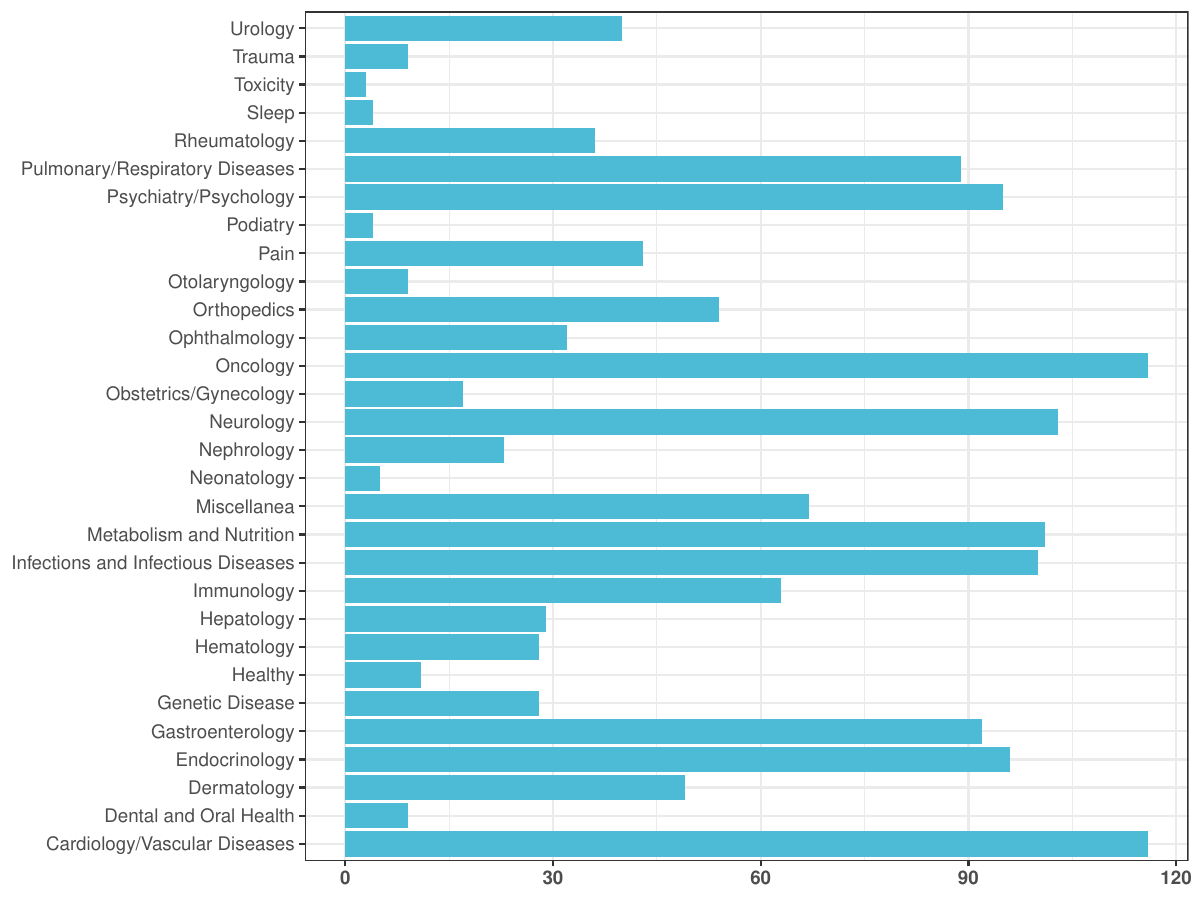}
\caption{Disease domains represented in EvidenceOutcomes.}
\label{fig:Disease}
\end{figure}

The disease domains featured in the 640 RCT abstracts were identified using a previously established ontology-based categorization of the `Conditions' listed on ClinicalTrials.gov, in line with their corresponding NCT numbers~\cite{Liu2022-dc}. Figure \ref{fig:Disease} shows the 30 disease domains. The top two disease domains were Cardiology/Vascular Diseases and Oncology. Sleep, Podiatry, and Toxicity had the lowest counts of PubMed abstracts among the two corpora.

\begin{figure}
\centering
\vspace{1em}
\includegraphics[width=.85\linewidth]{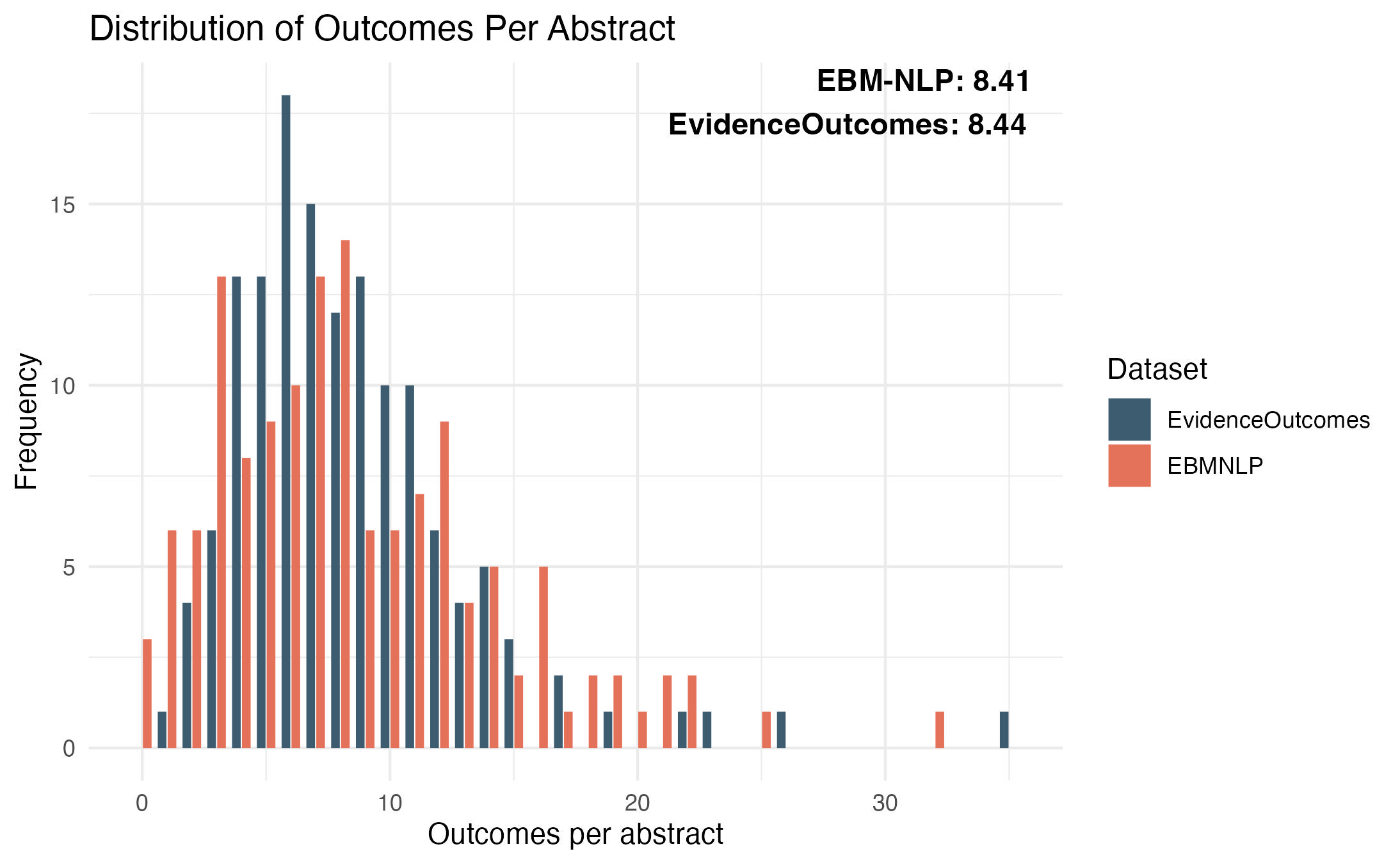}
\caption{Statistics of outcomes per abstract in EvidenceOutcomes and EBM-NLP.}
\label{fig:Statistics2}
\end{figure}

We conducted a detailed comparison of the common 140 abstracts from both EvidenceOutcomes and EBM-NLP datasets. The statistics of reported outcomes are detailed in Figure \ref{fig:Statistics2}. We found that the average number of outcomes annotated in EvidenceOutcomes was slightly higher than in EBM-NLP.

%

To provide a detailed breakdown, there were 606 overlapping outcomes, 650 outcomes unique to EvidenceOutcomes, and 695 unique to EBM-NLP. This difference highlighted that EvidenceOutcomes employs a distinct set of annotation guidelines compared to EBM-NLP. For example, in the abstract of PMID 19397502 (\url{https://pubmed.ncbi.nlm.nih.gov/19397502/}), EvidenceOutcomes provided a more focused outcome list, ``PI, MAP, toe temperature, PI, MAP changes, toe temperature changes, skin temperature'', while EBM-NLP annotated the whole sentence: ``PI in the toe, mean arterial pressure (MAP) and toe temperature, measures analysis of variance, clinically evident sympathectomy criteria (a 100\% increase in the PI, a 15\% decrease in MAP and a 1 degree C increase in toe temperature), McNemar test, photoplethysmography signals, PI, MAP and toe temperature, sympathectomy criteria, MAP, temperature changes, skin temperature, MAP.''

\section{Benchmark Results and Discussion}

To demonstrate the utility of EvidenceOutcomes, we propose one motivating use case for this annotated corpus that can be explored in future research efforts.

\subsection{Benchmark on EvidenceOutcomes}
BERT models \cite{Devlin2019-vb} have been widely used across many NLP tasks in the biomedical domain~\cite{Peng2019-Transfer}. Several encoder-based models have been further pre-trained on biomedical corpora, such as BioBERT \cite{lee2020biobert}, BlueBERT \cite{peng2019transfer}, PubMedBERT \cite{Gu2021-gu}, and BioELECTRA \cite{kanakarajan2021bioelectra}. In this work, we chose PubMedBERT as our baseline model because it demonstrates superior performance in previous PICO recognition tasks~\cite{Gu2021-gu, Zhang2024Span}.


To evaluate the baseline model, we used the following two datasets (Table~\ref{tab:data}). One is the EvidenceOutcomes dataset curated in this study, consisting of 500 RCT abstracts for training and 140 abstracts for testing. The other is the EBM-NLP in the study of Nye \textit{et al.}~\cite{Nye2018-ul}, which consists of 4,993 RCT abstracts describing RCTs. The training labels
were obtained via Amazon Mechanical Turk crowdsourcing. The test set was manually annotated by medical experts.

\begin{table}
\centering
\caption{Description of the datasets}
\label{tab:data}
\begin{tabular}{lrrrr}
\toprule
& \multicolumn{2}{c}{EvidenceOutcomes} & \multicolumn{2}{c}{EBM-NLP} \\
\cmidrule(rl){2-3}\cmidrule(rl){4-5}
& Training & Test & Training & Test\\
\midrule
Document & 500 & 140 & 4,547 & 186\\
Sentence & 8,063&2,021&50,106&2,089\\
Word&205,145&50,134&1,231,179&50,241\\
Entity&4,226&1,178&32,127&1,330\\
\bottomrule
\end{tabular}
\end{table}

To establish the baseline performance on the outcome element, we fine-tuned the PubMedBERT model using the training set of the EvidenceOutcomes and monitored the fine-tuning progress using the validation set, following the work of Nye \textit{et al.}~\cite{Nye2018-ul}. The model performance was evaluated using the test set of EvidenceOutcomes that contains labels from EBM-NLP, EvidenceOutcomes, and overlapping labels from both.

We implemented the CHILD-TUNING strategy \cite{Xu2021-hu} for fine-tuning, which updates only a subset of model parameters during training instead of all parameters in the regular fine-tuning. We used a learning rate of 3e-5, a batch size of 8, and default Adam Optimizer parameter values. 
The model was trained for 10 epochs, before the end of which the validation performance converged. The model checkpoint with the lowest validation loss was selected for evaluation. All the work was completed on two A6000 GPUs in the internal cluster.

We calculated the performance metrics (precision, recall, and F1 scores) at both the token level and the entity level. At the token level, a true positive case refers to an instance where the model correctly identifies the entity type that a token belongs to. At the entity level, a true positive refers to an instance where the model correctly identifies both the boundary and the category of an outcome entity. We used a Python package seqeval to calculate the scores~\cite{Seqeval2023-at}. We employed a bootstrap method, running 100 iterations to estimate the 95\% confidence interval and reported the intervals of these metrics.

The results on EvidenceOutcomes at the token and entity levels are shown in Table~\ref{tab:results_token} and Table~\ref{tab:results_entity}, respectively. Recall that the test data were annotated under two guidelines: EBM-NLP and EvidenceOutcomes. Hence, we reported the performance recognizing outcome entities under each annotation guideline. We also reported the performance on overlapping labels resulting from the two guidelines. 
The model exhibited the highest performance on instances with EvidenceOutcomes labels, achieving a precision of 0.77, a recall of 0.74, an F1-score of 0.76 at the token level, and a precision of 0.70, a recall of 0.67, an F1-score of 0.69 at the entity level. In contrast, the performance on instances with EBM-NLP labels is significantly lower, achieving a precision of 0.41, a recall of 0.30, an F1-score of 0.34 at the token level, and a precision of 0.27, a recall of 0.26, an F1-score of 0.26 at the entity level respectively. The better performance of the model on instances with EvidenceOutcomes labels could be attributed to the more comprehensive and detailed outcomes annotation guidelines used in the EvidenceOutcomes dataset. These results indicated a strong alignment between the model training and the characteristics specific to these outcome labels. On the other hand, the lower performance with EBM-NLP labels suggested that the outcomes distinguished by EBM-NLP were harder for the model to predict, and this further implied a discrepancy in the annotation guidelines between the EBM-NLP and EvidenceOutcomes datasets.  
%
\begin{table}
    \centering
    \caption{The performance of the model trained on 500 and tested on 140 EvidenceOutcomes abstracts at the token level. P - Precision, R - Recall, F1 - F1-score. Values in parentheses are 95\% confidence intervals}
    \label{tab:results_token}


\begin{tabular}{lccc}
\toprule
Label set & P & R & F1\\
\midrule
EBM-NLP Labels&0.405 (0.401, 0.409) &0.297 (0.293, 0.300)&0.339 (0.336, 0.342)  \\
EvidenceOutcomes Labels&0.772 (0.769, 0.775)&0.741 (0.737, 0.746)&0.756 (0.753, 0.758)\\
Overlapping Labels&0.455 (0.450, 0.461)&0.777 (0.771, 0.782)&0.572 (0.567, 0.577)\\
\bottomrule
\end{tabular}
\end{table}

\begin{table}
    \centering
    \caption{The performance of the model trained on 500 and tested on 140 EvidenceOutcomes abstracts at the entity level. P - Precision, R - Recall, F1 - F1-score. Values in parentheses are 95\% confidence intervals.}
    \label{tab:results_entity}
\begin{tabular}{lccc}
\toprule
Label set & P & R & F1\\
\midrule
EBM-NLP Labels&0.265 (0.261, 0.269)&0.258 (0.254, 0.261)&0.260 (0.257, 0.264)  \\
EvidenceOutcomes Labels&0.695 (0.692, 0.699)&0.673 (0.669, 0.677)&0.685 (0.681, 0.688) \\
Overlapping Labels&0.354 (0.349, 0.359)&0.711 (0.706, 0.717)&0.472 (0.467, 0.477)  \\
\bottomrule
\end{tabular}

\end{table}

\subsection{Benchmark on the combination of EBM-NLP and EvidenceOutcomes}

Next, we examined whether augmenting the existing PICO recognition dataset with EvidenceOutcomes will further improve the performance. We compared the model developed on EBM-NLP training data with the one developed on the combination of EBM-NLP and EvidenceOutcomes training data. Table~\ref{tab:resultsebm_token} and Table~\ref{tab:resultsebm_entity} present the evaluation results at the token and entity levels, respectively. Both models outperformed the existing benchmark presented by Nye et al~\cite{Nye2018-ul}. After combining EBM-NLP and EvidenceOutcomes training data, the model was further improved. At the entity level, the model achieved an F1-score of 0.44, increased from 0.39 on EBM-NLP, and an F1-score of 0.53, increased from 0.42 on EvidenceOutcomes, respectively.

\begin{table}
    \centering
    \caption{Performance of the model trained on the combined EBM-NLP training set and 500 EvidenceOutcomes abstracts, tested on EBM-NLP testing set and 140 EvidenceOutcomes abstracts at the token level. labels. P - Precision, R - Recall, F1 - F1-score.  Values in parentheses are 95\% confidence intervals
    }
    \label{tab:resultsebm_token}

\resizebox{\textwidth}{!}{%

\begin{tabular}{llccc}
\toprule
Training set & Test set & P & R & F1 \\
\midrule
EBM-NLP & EBM-NLP &0.711 (0.707, 0.716)&0.507 (0.503, 0.511) &0.592 (0.588, 0.595)\\
EBM-NLP + Evidence. & EBM-NLP&0.675 (0.670, 0.679)&0.612 (0.608, 0.617)&0.641 (0.637, 0.644)\\
EBM-NLP & Evidence. &0.498 (0.495, 0.501)&0.497 (0.494, 0.501)&0.493 (0.491, 0.496)\\
EBM-NLP + Evidence. & Evidence.&0.541 (0.538, 0.545)&0.673 (0.669, 0.677)&0.595 (0.592, 0.598)\\
\bottomrule
\end{tabular}}
\end{table}

\begin{table}
    \centering
    \caption{Performance of the model trained on the combined EBM-NLP training set and 500 EvidenceOutcomes abstracts, tested on EBM-NLP testing set and 140 EvidenceOutcomes abstracts at the entity level. P - Precision, R - Recall, F1 - F1-score.  Values in parentheses are 95\% confidence intervals
    }
    \label{tab:resultsebm_entity}
    \resizebox{\textwidth}{!}{%
\begin{tabular}{llccc}
\toprule
Training set & Test set & P & R & F1 \\
\midrule
EBM-NLP & EBM-NLP &0.501 (0.496, 0.505)&0.323 (0.319, 0.327)&0.392 (0.388, 0.396)\\
EBM-NLP + Evidence. & EBM-NLP&0.486 (0.481, 0.491)&0.397 (0.393, 0.402)&0.438 (0.434, 0.442)\\
EBM-NLP & Evidence. &0.485 (0.481, 0.489)&0.378 (0.375, 0.381)&0.424 (0.421, 0.427) \\
EBM-NLP + Evidence. & Evidence.&0.532 (0.528, 0.535)&0.530 (0.527, 0.533)&0.530 (0.527, 0.533)\\
\bottomrule
\end{tabular}}
\end{table}



\subsection{Limitations}

One limitation of the study is that the distribution of words, sentences, and outcomes in EvidenceOutcomes is not normalized, potentially impacting the generalizability of findings. The variation in abstract length and sentence structure across different studies may introduce biases, influencing the analysis and interpretation of results. Future research could explore normalization techniques to enhance the robustness of the dataset for more accurate and unbiased assessments.

Another limitation of the study stems from the discrepancies in the annotation schemes used by EvidenceOutcomes and EBM-NLP, though the scheme of EvidenceOutcomes is more comprehensive. These differences in annotation protocols hindered the possibility of carrying out a directly comparable experiment with EBM-NLP. To this end, we believe that EvidenceOutcomes offers a valuable addition to the existing resources.

\section{Conclusion}

Our work introduces an annotation guideline and a corpus for clinically meaningful outcomes in PubMed abstracts, as well as a trained NER model, which can be applied to downstream tasks. The tasks include aggregating RCT studies that investigate the same outcomes and assessing patient-centric outcomes, both aimed at enhancing evidence-based medicine practice.

\section*{Code Availability}

The codes can be found at: \url{https://github.com/ebmlab/EvidenceOutcomes}.

\section*{Acknowledgments}

This project was sponsored by the National Library of Medicine grant R01LM009886 and the National Center for Advancing Clinical and Translational Science award UL1TR001873.



\section*{Competing interests}

The authors declare no competing interests.

{
\bibliographystyle{unsrtnat}
\bibliography{sample}
}

\newpage



\section*{Supplementary material}

# Outcome Entity Extraction Annotation Guideline

Abigail M. Newbury[1], Betina Ross Idnay[1], Hao Liu[1], and Chunhua Weng[1,*]

[1]Columbia University, Department of Biomedical Informatics, New York, 10032, US
[*]corresponding author(s): Chunhua Weng (cw2384@cumc.columbia.edu)

## 1 Introduction

The goal of this guideline is to aid in creating a manually annotated corpus that will serve as the gold standard in training a machine learning (ML) model for outcome entity extraction. The outcome entity comes from the PICO framework (Population, Intervention, Comparison, Outcome) which assists in medical evidence formulation. The corpus for annotation comes from the abstracts of published Randomized Controlled Trial (RCT) studies. As this annotation focuses on outcome entity extraction, there will only be one type of tag (for outcome), though a sentence may contain multiple outcome entities. A detailed definition of an outcome entity is provided in the section below (Section 2), while full examples are provided in Section 4. Section 3 provides annotation logistics.

Because outcomes are defined by the study design, annotators will have access to the entire abstract and ClinicalTrials.gov specified Brief Summary and Primary and Secondary Outcomes Measures while annotating. This is described further in Section 3. Example annotations of full abstracts and their justification will be provided in Section 4 for this reason. However, for simplicity, in Section 2 only sentences will be provided, and the reader is asked to assume that highlighted entities are defined as outcomes per the study design.

## 2 Recognition of Outcome entity

### 2.1 What is an outcome?

The outcome of a study is the answer to the question: "What metrics are used to test the study hypothesis?" It constitutes the *measurable* result of a study. Thus, outcomes should only be annotated when they are *a part of the study design*. Outcomes are usually the dependent variable in a study, whose value depends on the treatment/intervention. Outcomes for RCT studies can include indicators of disease or symptom progression, recovery, safety, drug efficacy, and more. Outcome entities highlighted in text must have meaning outside of the context of the abstract, in other words, they must be able to stand alone. P-values, confidence intervals, and actual outcome observation values should never be included in outcome entities, and outcome entities should be contiguous (except for the exception of coordination ellipsis described in Section 2.3).

### 2.2 Modifiers

Modifiers are descriptive and *provide directionality* to the outcome entity but *should not be included in the outcome entity*.

(1)   The high-risk group displayed higher survival probability than the low-risk group.

The outcome entity is 'survival probability' while its modifier is 'higher'. Note how 'higher' is not included in the outcome entity, but instead serves to provide directionality to it. 'Survival probability' is a quantitative, measurable result, while 'higher' serves as a descriptor.

(2)   Mean change from baseline in mRNA levels was lower for the DTG arm as compared to placebo. (PMID: 25321146)

The modifier is 'lower' while the outcome entity is 'mean change from baseline in mRNA levels'. It is important to note that this outcome entity is structured differently from the example above, which just consisted of two nouns. The outcome entity includes 'mean' and the phrase 'change from baseline in'. The inclusion of words and phrases like these are discussed in Rules 12 and 8, respectively. Right now, this example serves to illustrate that outcome entities can take many forms so long as they are identified by the study design. So here, if the study design specifies measuring the change from baseline in mRNA levels it is highlighted, where 'mean' adds specificity.

(3) The <u>duration of a high viral load</u> was <u>shorter</u> with REGEN-COV than with placebo. (PMID: 34347950)

The 'duration of a high viral load' is an entity that a study could aim to measure. Modifiers regarding duration are typically 'long', 'slow', 'short', etc. which can be seen here with the modifier 'shorter'.

Words like increase, decrease, and reduction are typically used as modifiers and thus not included in the outcome entity (Example (33)). However, there are instances where these words should be included in the outcome as they are essential to the outcome concept. See Example (35) for an example of where 'improvement' should be included in the outcome.

### 2.3 Text Boundary

It is important to define the text boundary for an outcome entity. Note that as stated above, there are no restraints on the number of words that can be included in an outcome entity, or the overall structure of an outcome entity. However, conjunctions (and, or, as, but, etc.) *cannot be in an outcome entity*. An example is provided below.

(4) The <u>efficacy</u> and <u>safety</u> of nintedanib in Japanese patients were consistent with the overall INBUILD population. (PMID: 34564020)

Here, 'efficacy' and 'safety' are considered two separate outcome entities, and are not grouped together. There is a conjunction 'and' that separates them.

There are times when 'or' is not used as a conjunction but more so as part of a mathematical symbol. Consider these cases as an exception to the above rule. An example is provided below:

(5) <u>Grade 3 or worse adverse events</u> were more frequent in the control group.

Here, 'worse' has no meaning without its relationship with 'grade 3'. In a sense then, the 'or' acts to signify that they are looking for adverse events $\leq$ grade 3. Thus, in these specific cases 'or' may be included in the outcome entity.

Coordination ellipsis (CE) in text necessitates a specific way to mark the text boundary of an outcome entity. An example of this is as follows, which was taken from the Brat online platform.

(6) <u>glycopyrrolate</u> and <u>formoterol exposure</u> were decreased following administration (PMID: 28061907)

Here, 'glycopyrrolate exposure' and 'formoterol exposure' are the two outcomes in this sentence. To use the annotation tool with this type of annotation, one should highlight the contiguous entity 'formoterol exposure' as one, and then annotate 'glycopyrrolate' with an arrow to the overlapping annotation of just 'exposure'. The arrow is at 'exposure' because it comes last in the order of 'glycopyrrolate exposure'. Note that cases of coordination ellipsis are the only time non-contiguous outcomes should be annotated and the only time that outcome annotations are allowed to overlap.

### 2.4 Other PICO Elements

It is important to contextualize the outcome entity annotation process. There are additional PICO elements, which, for practical annotation purposes, are the P (population) and I (intervention) elements, and C (comparison) are typically grouped together in annotations). These elements *should not have any overlap* with the outcome entities. In other words, do not highlight any P or I elements as part of an outcome entity. The following provides an example of a sentence with the population element highlighted in blue and the intervention elements highlighted in green to make explicit the separation of these elements.

(7) <u>A1C</u> in <u>individuals with impaired glucose tolerance</u> was found to be lower with <u>inhaled insulin</u> vs. <u>placebo</u>. (PMID: 25271207)

Here, 'A1C' is the outcome entity, 'individuals with impaired glucose tolerance' is the population entity, and 'inhaled insulin' and 'placebo' are the two interventions.

### 2.5 Lists

When the term introducing the list is a clinical concept, and the introducing term and all elements of the list are a part of the study design, then the introducing term and its subsequent elements should all be highlighted as outcome entities. This is simply an example of redundant annotations which occur as per Rule 1.

(8) The counter-regulatory hormones glucagon, epinephrine, norepinephrine, growth hormone and cortisol were appropriately suppressed. (PMID: 25263215)

Here, 'counter-regulatory hormones' is a complete clinical concept, which can be broken down into the elements 'glucagon', 'epinephrine, norepinephrine', 'growth hormone' and 'cortisol', all of which are outcome metrics of the study. Thus, 'counter regulatory hormones' and all of the elements in the list are highlighted as outcome entities.

Sometimes the object introducing the list is too vague to act as an outcome entity on its own (recall the requirement that an outcome entity be able to "stand alone") and the elements are a part of the study design.

(9) Patients experienced the primary endpoints of nausea, flu-like symptoms, and headaches.

Here, 'primary endpoints' should not be marked as an outcome entity since that is both vague and not measurable. However, 'primary endpoints' serves to indicate to annotators that 'nausea', 'flu-like symptoms', and 'headaches' are all outcome metrics defined by the study design. Thus, each of 'nausea', 'flu-like symptoms', and 'headaches' are marked individually.

Additionally, sometimes in lists where the introducing term is an outcome entity, elements of the list were not specified as a part of the study design. This typically happens with patient-reported adverse events. When the object introducing the list is a clinical concept, and the elements of the list were not defined as part of the study design, the elements of the list should not be highlighted as outcome entities.

(10) Patients experienced adverse events such as nausea, flu-like symptoms, and headaches.

Here, 'adverse events' is a known, measurable outcome of many drug trials and thus is marked as the outcome entity. However, the elements of the list, namely, 'nausea', 'flu-like symptoms', and 'headaches' should not be highlighted as outcome entities as they are patient-reported and thus variable and unknown at the time of study design. They do not constitute as metrics used to test the study hypothesis.

(11) A majority of the participants were white, female, and above the age of 30.

Here, it is important to note that baseline characteristics of a study (that are not outcomes of the study) act as covariates or classification variables. Thus, none of the entities mentioned in the above sentence are outcomes as defined by the study design.

Similarly, sometimes multivariate analysis is used to find association between outcome entities and covariates. In these cases, only the outcome entities should be highlighted.

(12) In a multivariate analysis, high plasma ribavirin concentration, low baseline hemoglobin, HCV genotype 1b, and IL28B genotype CC were associated with higher SVR12. (PMID: 26650626)

Here, the covariates are 'plasma ribavirin concentration, 'baseline hemoglobin', 'HCV genotype 1b', and 'IL28B genotype CC', which are all associated with the outcome entity 'SVR12', but should not themselves be highlighted as outcome entities.

## 2.6 Temporal Component

Outcome entities will occasionally have a temporal component that is essential to the meaning of the outcome. There are certain indicators to use to determine if the temporal component adjacent to an outcome is essential to that outcome and should be highlighted. The indicators are as follows.

1. The temporal component is *explicitly stated by the study design.* This means that the temporal component is either, i) stated as part of the defined outcome in sections prior to the results and conclusions sections in the abstract or ii) stated as part of the defined outcome in the ClinicalTrials.gov file Primary or Secondary Outcomes Measures. Example (32) provides an example of this where 'adjusted mean change in HIV-1 RNA at day 8' is the outcome.

2. The temporal component is used as an adjective preceding the rest of the outcome entity. Typically, adjectival use of temporal components indicates that the inclusion of the temporal component adds clinical significance and meaning to the outcome. An example of this is as follows.

(13) Changes from baseline with dapagliflozin 10 mg by day 7 were -2.29 mmol/L for 24-h daily average blood glucose. (PMID: 25271207)

Here, '24-h' is the temporal component being used adjectivally preceding the rest of the outcome, namely, 'daily average blood glucose'. Thus, it is included as part of the outcome entity.

## 2.7 Additional Rules

**Rule 1.** Outcomes should be highlighted each time they are mentioned as an outcome

**Rule 2.** Outcomes beyond the scope of the current study should not be annotated as outcome entities

**Rule 3.** Articles preceding outcome entities should never be included in outcome entities

(14)  The <u>blood glucose level</u> increased.

**Rule 4.** Always highlight the outcome entities: tolerability, efficacy, and safety. This includes all derivations of these words, but not any words that come after or before.

(15)  <u>well-tolerated</u>, <u>tolerated</u>, <u>tolerability</u>

(16)  <u>efficacious</u>

(17)  <u>safety</u> profile

(18)  <u>efficacy</u> parameter

(19)  short-term <u>tolerability</u> variables

**Rule 5.** Always highlight the outcome entity: adverse event(s). This includes any subsets, but not anything else that comes after/before.

(20)  <u>treatment-emergent adverse events</u>

(21)  <u>grade 3/4 adverse events</u>

(22)  <u>treatment-emergent adverse event</u> rates

(23)  rates of <u>mild gastrointestinal adverse events</u>

**Rule 6.** Never highlight feasibility, effectiveness, superiority, or noninferiority (or their derivations) as stand-alone outcomes. They should not be treated the same as those mentioned in Rule 4.

**Rule 7.** Do not highlight discontinuations or deaths as outcomes unless *explicitly stated by study design*

**Rule 8.** Only highlight 'from baseline' if 'changes in', 'differences in', 'improvement in', or a similar entity is also included in the outcome

(24)  At month 12, there was no significant difference in the <u>change in serum Cr level from baseline</u>. (PMID: 26894653)

(25)  Once-daily 5% MTF resulted in <u>differences in TAHC from baseline</u> by 23.9 ± 2.1 hairs/cm at week 24. (PMID: 27391640)

**Rule 9.** Numeric values can be included in outcome entities if part of study design

(26)  <u>recovery time to glucose level ≤ 3.9 mmol/l</u> (PMID: 25263215)

**Rule 10.** Parentheses (and the words/numbers that they encapsulate) adjacent to an outcome entity that are a part of the outcome entity (and not its observed value, CI, etc.) can be highlighted as a part of that outcome entity

(27)  <u>recovery time to glucose level ≤3.9 mmol/l (≤70 mg/dl)</u> (PMID: 25263215)

**Rule 11.** An abbreviation of an outcome entity in parentheses immediately after the outcome entity should be highlighted separately as an outcome entity (redundancy)

(28)  <u>adverse events</u> (<u>AEs</u>)

**Rule 12.** Always try to include as much specificity in the outcome entity annotation as possible. If specificity is provided by coordinate adjectives, commas can be included in the outcome entity

(29)  <u>muscle invasive, p53-like bladder cancer progression</u>

**Rule 13.** Phrases like 'need for', 'measures of', 'risk of', 'incidence of', 'occurrence of' generally shouldn't be highlighted as part of the outcome entity

**Rule 14.** Highlight outcomes at baseline if mentioned

(30)  <u>mean baseline HbA1c</u>

(31)  <u>pain score at baseline</u>

## 3  Logistics

The text corpus will be uploaded to the online platform called Brat. For abstracts with clear section labels, the user should annotate the results and conclusion section. If there are no clear labels, the user should use their best judgment to annotate the results and conclusion sections. The ClinicalTrials.gov fields Brief Summary, Primary Outcome Measures and Secondary Outcome Measures will be provided for each RCT abstract in a .txt file. Example .txt files are shown in Section 4.

The protocol for referencing ClinicalTrials.gov .txt files is as follows. Annotators should annotate outcome entities as identified by the content of the abstract. In cases when the annotator is unclear if a term is an outcome entity, the annotator should consult the ClinicalTrials.gov .txt file. This protocol decreases the burden of cross-referencing on annotators and serves as a small safeguard against a potential incompleteness of outcomes listed on ClinicalTrials.gov. The inclusion of the reference ClinicalTrials.gov study design should make the annotation process clearer when there is uncertainty.

## 4  Full Abstracts

Note that as per Section 3, only the results and conclusions sections will be annotated. Additionally note that as per the protocol in Section 3, outcomes will be inferred from the context of the abstract unless otherwise stated.

(32)    PMID: 25321146, NCTID: NCT01568892

Brief Summary: Study ING116529 is a multicepter, randomized, study with an initial 7 day placebo- controlled, functional monotherapy phase to quantify the antiviral activity attributable to dolutegravir (DTG) in HIV-1 infected, ART-experienced adults who are experiencing virological failure on an Integrass inhibitor containing regimen (current RAL or ELV failures), with evidence of genotypic resistance to RAL or ELV at study entry. Thirty subjects will be randomized (1:1) to receive either DTG 50mg BID (Arm A) or Placebo (Arm B) with the current failing regimen for 7 days (RAL or ELV should be discontinued prior to dosing with DIG). At Day 8, subjects from both arms will enter an open label phase and receive open label DTG 50mg BID with an optimized background regimen containing at least one fully active drug.

Primary Outcome Measures:

1. Mean Change From Baseline in Plasma Human Immunodeficiency Virus Type 1 (HIV-1) Ribonucleic Acid (RNA) at Day 8

Secondary Outcome Measures:

1. Absolute Values in Plasma HIV-1 RNA Over Time
2. Mean Change From Baseline in Plasma HIV-1 RNA Over Time
3. Number of Participants With Plasma HIV-1 RNA <50 c/mL Over Time
4. Number of Participants With Plasma HIV-1 RNA <400 c/mL Over Time
5. Absolute Values in Cluster of Differentiation 4+ (CD4+) Cell Counts Over Time
6. Median Change From Baseline in CD4+ Cell Counts Over Time
7. Absolute Values in Cluster of Differentiation 8+ (CD8+) Cell Counts Over Time
8. Median Change From Baseline in CD8+ Cell Counts Over Time
9. Number of Participants With the Indicated Type of HIV-1 Disease Progression (Acquired Immunodeficiency Syndrome [AIDS] or Death [DM])
10. Number of Participants With Any Adverse Event (Serious and Non-serious) of the Indicated Grade
11. Number of Participants With the Maximum Post-Baseline-emergent Clinical Chemistry Toxicities of the Indicated Grade
12. Number of Participants With the Maximum Post-Baseline-emergent Hematology Toxicities of the Indicated Grade
13. AUC(0-tau) of DTG
14. Cmax of DTG
15. Plasma DTG Pre-dose Concentration (CO) at Day 8, Day 28, and Week 24; and Average DTG CO (CO An) at Week 24
16. Number of Participants With the Indicated Treatment-emergent Integrase (IN) Mutations Detected at the Time of Defined Virologic Failure (PDVF), as a Measure of Genotypic Resistance
17. Number of Participants With the Indicated Fold Increase in Fold Change (FC) in the 50% Inhibitory Concentration Relative to Wild-type Virus for DTG (i.e. PDVF FC/Baseline FC Ratio) at the Time of PDVF, as a Measure of Phenotypic Resistance
18. Number of Participants Who Discontinued Study Treatment Due to AEs
19. Number of Participants With Clinically Significant Electrocardiogram (ECG) Findings
20. Change From Baseline in Systolic Blood Pressure (SBP) and Diastolic Blood Pressure (DBP)
21. Change From Baseline in Heart Rate
22. Change From Baseline in Albumin Level
23. Change From Baseline in Alkaline Phosphatase (ALP), Alanine Aminotransferase (ALT), Aspartate Aminotransferase (AST) and Creatine Kinase
24. Change From Baseline in Total Bilirubin (T. 11,4) and Creatinine Levels
25. Change From Baseline in Cholesterol, Chloride, Carbon Dioxide (CO2)/Bicarbonate (HCO3), Glucose, High Density Lipoprotein Cholesterol, Potassium, Low Density Lipoprotein (LDL) Cholesterol, Sodium, Phosphorus, Triglycerides and Urea/Blood Urea Nitrogen
26. Change From Baseline in Creatinine Clearance
27. Change From Baseline in Lipase Levels
28. Change From Baseline in Basophils, Eosinophils, Lymphocytes, Monocytes, Total Neutrophils, Platelet and White Blood Cell (NBC) Count
29. Change From Baseline in Hemoglobin Level
30. Change From Baseline in Hematocrit Level
31. Change From Baseline in Mean Corpuscle Volume
32. Change From Baseline in Red Blood Cell Count

| | |
|---|---|
| 1 | Dolutegravir versus placebo in subjects harbouring HIV-1 with integrase inhibitor resistance associated substitutions: 48-week results from VIKING-4, a randomized study. |
| 3 | BACKGROUND : The Phase III VIKING-3 study demonstrated that dolutegravir (DTG) 50 mg twice daily was efficacious in antiretroviral therapy (ART)-experienced subjects harbouring raltegravir- and/or elvitegravir-resistant HIV-1. |
| 4 | VIKING-4 (ING116529) included a placebo-controlled 7-day monotherapy phase to demonstrate that short-term antiviral activity was attributable to DTG. |
| 6 | METHODS : VIKING-4 is a Phase III randomized, double-blind study in therapy-experienced adults with integrase inhibitor (INI)-resistant virus randomized to DTG 50 mg twice daily or placebo while continuing their failing regimen (without raltegravir or elvitegravir) for 7 days (clinicaltrials.gov identifier NCT01568892). |
| 7 | At day 8, all subjects switched to open-label DTG 50 mg twice daily and optimized background therapy including ≥1 fully active drug. |
| 8 | The primary end point was change from baseline in plasma HIV-1 RNA at day 8. |
| 10 | RESULTS : The study population (n=30) was highly ART-experienced with advanced HIV disease. |
| 11 | Patients had extensive baseline resistance to all approved antiretroviral classes. |
| 12 | Adjusted mean change in HIV-1 RNA at day 8 was -1.06 log10 copies/ml for the DTG arm and 0.10 log10 copies/ml for the placebo arm (treatment difference -1.16 log10 copies/ml [-1.52, -0.80]; P<0.001). |
| 13 | Overall, 47% and 57% of subjects had plasma HIV-1 RNA <50 and <400 copies/ml at week 24, and 40% and 53% at week 48, respectively. |
| 14 | No discontinuations due to drug-related adverse events occurred in the study. |
| 16 | CONCLUSIONS : The observed day 8 antiviral activity in this highly treatment-experienced population with INI-resistant HIV-1 was attributable to DTG. |
| 17 | Longer-term efficacy (after considering baseline ART resistance) and safety during the open-label phase were in-line with the results of the larger VIKING-3 study. |

Lines 10 and 11 provide baseline characteristics about the study population and thus nothing in these sentences is highlighted as an outcome entity.

Line 12 consists of the outcome entity 'adjusted mean change in HIV-1 RNA at day 8'. From the context of the abstract (especially Sentence 7), the primary outcome involves measuring the change in HIV-1 RNA where 'day 8' and 'change in' and are defined as a part of the outcome entity in Sentence 7, satisfying Section 2.6 and Rule 8, respectively. 'Adjusted' serves to increase the specificity of the outcome entity and thus is included as per Rule 12. If there was any lack of clarity, this outcome entity can be further confirmed by Primary Outcome Measure 1 in the ClinicalTrials.gov file. This process exemplifies the protocol of first inferring outcomes from the semantics of the abstract and if there is lack of clarity, subsequently checking the ClinicalTrials.gov file. 'Treatment difference' in line 12 is not highlighted as an outcome entity as it is not stand alone (i.e. treatment difference *in what*).

In line 13, the two outcome entities 'plasma HIV-1 RNA <50 copies/ml' and 'plasma HIV-1 RNA <400 copies/ml' are highlighted. Weeks 24 and 48 are not *explicitly stated by the study design*, thus they are not essential components of the outcome and are not included. The inclusion of the values (copies/ml) is permitted per Rule 9.

In line 14, the question of whether to include 'discontinuations due to' is addressed by Rule 7 which requires 'discontinuations due to' to be *explicitly stated by the study design*. While it is not mentioned in the background or methods sections, 'number of participants who discontinued study treatment due to AEs' is mentioned in Secondary Outcome Measure 18 in the ClinicalTrials.gov file. Thus, 'discontinued due to' is included in the outcome entity.

In line 16, 'day 8 antiviral activity' is highlighted as an outcome entity as it is a more general term for the outcome highlighted in line 12. Lastly, 'efficacy' and 'safety' are highlighted in line 17 per Rule 4.

(33)    PMID: 23715753, NCTID: NCT01068665

> Brief Summary: This trial is conducted in South Africa, Europe and North America. The aim of this trial is to compare efficacy and safety of NN1250 (insulin demystes (IDeg)) with insulin glargine (IGlar), as add—on to subject's ongoing treatment with metformin and/or qipeptyl peptidase 4 (DPP-4) inhibitors, in patients with type 2 diabetes being treated with oral anti—diabetic drugs (OADs) qualifying for intensified treatment.
>
> Primary Outcome Measures: 1. Change in Glycosylated Haemoglobin (HbA1c)
>
> Secondary Outcome Measures: 1. Change in Fasting Plasma Glucose (FPG)

```
1  Low-volume insulin degludec 200 units/ml once daily improves glycemic control similarly to insulin glargine with a low risk of hypoglycemia in insulin-naive patients with type
   2 diabetes: a 26-week, randomized, controlled, multinational, treat-to-target trial: the BEGIN LOW VOLUME trial.

3  OBJECTIVE : The 200 units/mL formulation of insulin degludec (IDeg 200 units/mL) contains equal units of insulin in half the volume compared with the 100 units/mL
   formulation.
4  We compared the efficacy and safety of IDeg 200 units/mL once daily with 100 units/mL insulin glargine (IGlar) in insulin-naïve subjects with type 2 diabetes (T2DM)
   inadequately controlled with oral antidiabetic drugs.

6  METHODS : In this 26-week, open-label, treat-to-target trial, subjects (n = 457; mean HbA1c 8.3% [67 mmol/mol], BMI 32.4 kg/m(2), and fasting plasma glucose [FPG] 9.6
   mmol/L [173.2 mg/dL]) were randomized to IDeg 200 units/mL or IGlar, both given once daily in combination with metformin with or without a dipeptidyl peptidase-4 inhibitor.
7  Basal insulin was initiated at 10 units/day and titrated weekly to an FPG target of <5 mmol/L (<90 mg/dL) according to mean prebreakfast self-measured blood glucose
   values from the preceding 3 days.

                            [Outcome]
9  RESULTS : By 26 weeks, IDeg reduced  HbA1c  by 1.30% and was not inferior to IGlar.
               [Outcome]
10 Mean observed FPG reductions were significantly greater with IDeg than IGlar (-3.7 vs. -3.4 mmol/L [-67 vs. -61 mg/dL]; estimated treatment difference: -0.42 [95% CI -0.78
   to -0.06], P = 0.02).
                                [Outcome]
11 Despite this difference, rates of overall confirmed hypoglycemia were not higher with IDeg than with IGlar (1.22 and 1.42 episodes/patient-year, respectively), as were
                      [Outcome]
   rates of nocturnal confirmed hypoglycemia (0.18 and 0.28 episodes/patient-year, respectively).
            [Outcome]
12 Mean daily basal insulin dose was significantly lower by 11% with IDeg 200 units/mL compared with IGlar.
               [Outcome]          [Outcome]
13 IDeg was well-tolerated, and the rate of treatment-emergent adverse events was similar across groups.

                                                                   [Outcome]
15 CONCLUSIONS : In this treat-to-target trial in insulin-naïve patients with T2DM, IDeg 200 units/mL improved glycemic control similarly to IGlar with a low risk of
         [Outcome]
   hypoglycemia.
```

In line 9, 'HbA1c' is highlighted as an outcome entity. The word 'reduced' in front of 'HbA1c' acts as a modifier and thus is not included in the outcome entity.

In line 10, 'mean observed FPG' is highlighted as the outcome entity. Here, 'reductions' is treated as a modifier and thus is not included in the outcome entity (it does not appear that they are interested in measuring 'reductions' specifically, just 'change in'). The adjectives 'mean observed' are included in the outcome entity to add additional specificity as per Rule 12.

In line 11, 'rates of overall confirmed hypoglycemia' and 'rates of nocturnal confirmed hypoglycemia' are highlighted as outcome entities. 'Rates of' adds specificity to both entities and is included per Rule 12.

'Mean daily basal insulin dose' is highlighted as an outcome in line 12, where 'mean daily basal' adds specificity (Rule 12).

In line 13, 'well-tolerated' is highlighted per Rule 4, and 'treatment-emergent adverse events' is highlighted per Rule 5 where 'treatment-emergent' is included as it subsets 'adverse events' and 'rate of' is not included.

In line 15, 'glycemic control' and 'hypoglycemia' are highlighted as they act as general, concluding terms to refer to the more specific outcome entities mentioned in the results section. Clearly, the ClinicalTrials.gov file corresponding to this abstract is nowhere near as comprehensive as in the previous example. However, this example shows how inferences can be made from the abstract itself.

(34) PMID: 28343225, NCTID: NCT01381094

Brief Summary: The purpose of this study is to evaluate the dose response (efficacy), pharmacodynamic response, pharmacokinetics, safety, and tolerability of orally administered AKB-6548 in pre-dialysis participants with anemia with repeat dosing for 42 days.

Primary Outcome Measures:

1. Absolute Change From Baseline in Hemoglobin (Hgb) to End of Treatment (Week 6)

Secondary Outcome Measures:

1. Change From Baseline in HO at Week 1, Week 2, Week 4, Week 6, and Follow-up Visit (up to Week 8)
2. Change From Baseline in Hematocrit (HCT) at Week 1, Week 2, Week 4, Week 6, and Follow-up Visit (up to Week 8)
3. Change From Baseline in Red Blood Cell (RBC) Count at Week 1, Week 2, Week 4, Week 6, and Follow-up Visit (up to Week 8)
4. Change From Baseline in Absolute Reticulocyte Count at Week 1, Week 2, Week 4, Week 6, and Follow-up Visit (up to Week 8)
5. Change From Baseline in Reticulocyte Hgb Content at Week 6
6. Maximum Change From Baseline in 110
7. Maximum Change From Baseline in HCT
8. Maximum Change From Baseline in RBC Count
9. Maximum Change in Absolute Reticulocyte Count From Baseline
10. Number of Participants With Absolute Change From Baseline in Hgb a 0.4, 0.6, 0.8, and 1.0 g/dL at the End of Dosing Period
11. Number of Participants With Change From Baseline in Ho a5.0, 7.5, and 10.0% by the End of Dosing Period
12. Number of Participants With Change From Baseline in HCT a5.0, 7.5, and 10.0% by the End of Dosing Period
13. Number of Participants With Change From Baseline in RBC Count a5.0, 7.5, and 10.0% by the End of Dosing Period
14. Number of Participants With Change From Baseline in Reticulocyte Count a6000, 12000, and 18000 Cells/uL by the End of Dosing Period
15. Change From Baseline in Total Iron at Week 2, Week 4, Week 6, and Follow-up Visit (up to Week 8)
16. Change From Baseline in Unsaturated Iron Binding Capacity at Week 2, Week 4, Week 6, and Follow-up Visit (up to Week 8)
17. Change From Baseline in Iron Saturation at Week 2, Week 4, Week 6, and Follow-up Visit (up to Week 8)
18. Change From Baseline in Total Iron Binding Capacity (TIBC) at Week 2, Week 4, Week 6, and Follow-up Visit (up to Week 8)
19. Change From Baseline in Ferritin at Week 2, Week 4, Week 6, and Follow-up Visit (up to Week 8)
20. Change From Baseline in Erythropoietin at Week 2, Week 4, Week 6, and Follow-up Visit (up to Week 8)
21. Change From Baseline in tieRciain at Week 6
22. Mean Plasma Vadadustat Concentrations on Week 2 and Week 4
23. Mean Plasma Vadadustat Acyl-Glucuronide Concentrations on Week 2 and Week 4
24. Number of Participants Treatment-emergent Adverse Events (TEAEs) and Serious Adverse Events (SAEs)
25. Number of Participants With Clinically Significant Changes From Baseline in Vital Signs Parameter
26. Number of Participants With Clinically Significant Abnormal 12-Electrocardiogram (ECG) Findings
27. Mean Change From Baseline in PR Interval, QT Interval, QRS Interval, and QT Corrected (OTC) Interval
28. Number of Participants With Clinically Significant Changes From Baseline in Laboratory Parameter Values

---

1  Clinical Trial of Vadadustat in Patients with Anemia Secondary to Stage 3 or 4 Chronic Kidney Disease.

3  BACKGROUND : Therapeutic options for the treatment of anemia secondary to chronic kidney disease (CKD) remain limited.
4  Vadadustat (AKB-6548) is an oral hypoxia-inducible factor prolyl-hydroxylase domain (HIF-PHD) inhibitor that is being investigated for the treatment of anemia secondary to CKD.

6  METHODS : A phase 2a, multicenter, randomized, double-blind, placebo-controlled, dose-ranging trial (NCT01381094) was undertaken in adults with anemia secondary to CKD stage 3 or 4.
7  Eligible subjects were evenly randomized to 5 groups: 240, 370, 500, or
8  630 mg of once-daily oral vadadustat or placebo for 6 weeks.
9  All subjects received low-dose supplemental oral iron (50 mg daily).
10 The primary endpoint was the mean absolute change in hemoglobin (Hb) from baseline to the end of treatment.
11 Secondary endpoints included iron indices, safety, and tolerability.

13 RESULTS : Ninety-three subjects were randomized.
14 Compared with placebo, vadadustat significantly increased Hb after 6 weeks in a dose-dependent manner (analysis of variance; p < 0.0001).
15 Vadadustat increased the total iron-binding capacity and decreased concentrations of ferritin and hepcidin.
16 The proportion of subjects with at least 1 treatment-emergent adverse event was similar between vadadustat- and placebo-treated groups.
17 No significant changes in blood pressure, vascular endothelial growth factor, C-reactive protein, or total cholesterol were observed.
18 Limitations of this study included its small sample size and short treatment duration.
20 CONCLUSIONS : Vadadustat increased Hb levels and improved biomarkers of iron mobilization and utilization in patients with anemia secondary to stage 3 or 4 CKD.
21 Global multicenter, randomized phase 3 trials are ongoing in non-dialysis-dependent and dialysis-dependent patients.
23 © 2017 The Author(s) Published by S. Karger AG, Basel.

Sentence 13 specifies details of the trial procedure and thus nothing is highlighted as an outcome. In sentence 14, 'Hb after 6 weeks' is highlighted as the outcome entity. 'After 6 weeks' is included because of its explicit mention in Primary Outcome Measure 1 as well as Secondary Outcome Measure 1. In line 15, 'total iron-binding capacity' is highlighted. Additionally, 'concentrations of ferritin' and 'concentrations of hepcidin' are highlighted using the coordination ellipsis highlighting procedure described in Section 2.3. The outcome 'treatment-emergent adverse event' is highlighted in line 16 as per Rule 5. In line 17, 'blood pressure', 'vascular endothelial growth factor', 'C-reactive protein' and 'total cholesterol' are highlighted as outcome entities. Note that 'changes in' is not included in the outcomes as it is not *explicitly stated by the study design*. Lastly, in sentence 20, 'Hb levels' is highlighted as it is a restatement of the outcome in line 14. Additionally, 'biomarkers of iron mobilization' and 'biomarkers of utilization' are highlighted via the coordination ellipsis highlighting procedure.

(35)    PMID: 35286843, NCTID:NCT04317040

> Brief Summary: This study evaluates the efficacy and safety of CD24Fc (MK-7110) in hospitalized adult participants who are diagnosed with coronavirus disease 2019 (COVID-19) and receiving oxygen support.
>
> The primary hypothesis of the study is clinical improvement in the experimental group versus the control group.
>
> Primary Outcome Measures:
>
> 1. Time to Improvement in Coronavirus Disease 2019 (COVID-19) Clinical Status
> 2. Number of Participants Who Experience an Adverse Event (AE)
>
> Secondary Outcome Measures:
>
> 1. Percentage of Participants Who Died or Had Respiratory Failure (RF)
> 2. Time to Disease Progression in Clinical Status of COVID-19
> 3. Number of Participants Who Died Due to Any Cause
> 4. Rate of Clinical Relapse
> 5. Conversion Rate of COVID-19 Clinical Status
> 6. Time to Hospital Discharge
> 7. Duration of MV
> 8. Duration of Pressors
> 9. Duration of ECMO
> 10. Duration of High Flow Oxygen Therapy
> 11. Length of Hospital Stay
> 12. Change From Baseline in Absolute Lymphocyte Count
> 13. Change From Baseline in D—Dimer Concentration

```
 7  METHODS : We conducted a randomised, double-blind, placebo-controlled, phase 3 study at nine medical centres in the USA.
 8  Hospitalised patients (age ≥18 years) with confirmed SARS-CoV-2 infection who were receiving oxygen support and standard of care were randomly assigned (1:1) by site-stratified block randomisation to receive a single intravenous
    infusion of CD24Fc 480 mg or placebo.
 9  The study funder, investigators, and patients were masked to treatment group assignment.
10  The primary endpoint was time to clinical improvement over 28 days, defined as time that elapsed between a baseline National Institute of Allergy and Infectious Diseases ordinal scale score of 2-4 and reaching a score of 5 or higher or
    hospital discharge.
11  The prespecified primary interim analysis was done when 146 participants reached the time to clinical improvement endpoint.
12  Efficacy was assessed in the intention-to-treat population.
13  Safety was assessed in the as-treated population.
14  This study is registered with ClinicalTrials.gov, NCT04317040.

16  FINDINGS : Between April 24 and Sept 22, 2020, 243 hospitalised patients were assessed for eligibility and 234 were enrolled and randomly assigned to receive CD24Fc (n=116) or placebo (n=118).
17  The prespecified interim analysis was done when 146 participants reached the time to clinical improvement endpoint among 197 randomised participants.
18  In the interim analysis, the [Outcome]28-day clinical improvement rate[/Outcome] was 82% (81 of 99) for CD24Fc versus 66% (65 of 98) for placebo; [Outcome]median time to clinical improvement[/Outcome] was 6·0 days (95% CI 5·0-8·0) in the CD24Fc group versus 10·0 days
    (7·0-15·0) in the placebo group (hazard ratio [HR] 1·61, 95% CI 1·16-2·23; log-rank p=0·0028, which crossed the prespecified efficacy boundary [α=0·0147]).
19  37 participants were randomly assigned after the interim analysis data cutoff date; among the 234 randomised participants, [Outcome]median time to clinical improvement[/Outcome] was 6·0 days (95% CI 5·0-9·0) in the CD24Fc group versus 10·5 days
    (7·0-15·0) in the placebo group (HR 1·40, 95% CI 1·02-1·92; log-rank p=0·037).
20  The proportion of participants with [Outcome]disease progression[/Outcome] within 28 days was 19% (22 of 116) in the CD24Fc group versus 31% (36 of 118) in the placebo group (HR 0·56, 95% CI 0·33-0·95; unadjusted p=0·031).
21  The incidences of [Outcome]adverse events[/Outcome] and [Outcome]serious adverse events[/Outcome] were similar in both groups.
22  No [Outcome]treatment-related adverse events[/Outcome] were observed.

24  INTERPRETATION : CD24Fc is generally well [Outcome]tolerated[/Outcome] and accelerates [Outcome]clinical improvement[/Outcome] of hospitalised patients with COVID-19 who are receiving oxygen support.
25  These data suggest that targeting inflammation in response to tissue injuries might provide a therapeutic option for patients hospitalised with COVID-19.

27  FUNDING : Merck & Co, National Cancer Institute, OncoImmune.

29  Copyright © 2022 Elsevier Ltd. All rights reserved.
```

Lines 16 and 17 describe the trial design.

In line 18, '28-day clinical improvement rate' and 'median time to clinical improvement' are highlighted as outcomes, as indicated by line 10. Note here that '28-day' is an adjective to 'clinical improvement rate' and thus is included as per Section 2.6.

In line 19, 'median time to clinical improvement' is again highlighted as an outcome (redundancy).

In line 20, 'disease progression' is highlighted as an outcome but 'within 28 days' is excluded as it is not *explicitly stated by the study design*.

In line 21, 'adverse events' and 'serious adverse events' are highlighted as outcomes as per Rule 5 ('serious' serves to subset 'adverse events').

Similarly, 'treatment-related adverse events' is highlighted in line 22.

Lastly, in line 24, 'tolerated' is highlighted as per Rule 12 and 'clinical improvement' is highlighted because it is a redundant mention of an outcome (Rule 1).



\end{document}


\let\linenumbers\nolinenumbers\nolinenumbers
\maketitle

\section{Extra information}

\paragraph{1(a) Dataset documentation and intended uses}
The fundamental process of evidence extraction and synthesis in the practice of evidence-based medicine involves extracting PICO (Population, Intervention, Comparison, and Outcome) elements from biomedical literature. However, Outcomes, being the most complex elements, are often overlooked or neglected. To address this issue, we developed a robust annotation guideline for extracting clinically meaningful outcomes from text through iteration and discussion with clinicians and Natural Language Processing experts.

Three independent annotators annotated the Results and Conclusions sections of a randomly selected sample of 500 PubMed abstracts and an additional 140 PubMed abstracts from the existing EBM-NLP corpus. This resulted in a dataset, EvidenceOutcomes, with high-quality annotations of an inter-rater agreement of 0.76.

\paragraph{1(b) URL to website/platform hosting the dataset.}
\url{https://github.com/ebmlab/EvidenceOutcomes}

\paragraph{1(c) URL to Croissant metadata record.}
\url{https://github.com/ebmlab/EvidenceOutcomes/blob/main/EvidenceOutcomes_meta_data_mlc.json}

\paragraph{1(d) Author statement}
We hereby affirm that we are the authors and proprietors of the dataset titled EvidenceOutcomes ("the Dataset"). We accept full responsibility for the content of the Dataset and confirm that I have legally obtained this data. Furthermore, we warrant that the Dataset does not infringe on any copyrights, trademarks, patents, or other proprietary rights, privacy rights, or any other rights of third parties.

We confirm that the Dataset is provided under the terms of the MIT license, which permits distribution, modification, commercial use, and private use with minimal restrictions. Users must include the original copyright and license notice in any copy of the software or substantial portions of the software. The Dataset is provided "as is", without warranty of any kind. The license explicitly states that the author or copyright holder is not liable for any claim, damages, or other liability, whether in an action of contract, tort, or otherwise, arising from, out of, or in connection with the software or the use or other dealings in the software. 

\paragraph{1(e) Hosting, licensing, and maintenance plan}
EvidenceOutcimes ("the Dataset") is stored and backed up in a local lab server managed by the Department of Biomedical Informatics at Columbia University Irving Medical Center. The latest version is released on GitHub platform. Distributed under the terms of the MIT license, the Dataset is free and open-source.

To ensure data quality, we developed a robust annotation guideline that is published along with the Dataset. To avoid potential individual bias, at least two lab members participated in annotation for each data point strictly following the guideline. 

We used the MIT license which can be viewed at \url{https://opensource.org/license/mit}.

\paragraph{2(a) Links to access the dataset and its metadata} \url{https://github.com/ebmlab/EvidenceOutcomes}

\paragraph{2(b) Data format} We used BRAT (Brat Rapid Annotation Tool, \url{https://brat.nlplab.org/}), which is a web-based annotation tool designed for collaborative text annotation tasks, particularly in the field of NLP and machine learning, to create the annotations.

The free text of the selected PubMed abstract, brat configuration files, and annotated data files are all available at \url{https://github.com/ebmlab/EvidenceOutcomes}.

\paragraph{2(c) Long-term preservation} The GitHub page is publicly accessible. The duration of preservation and sharing of the data will be a minimum of 5 years after the funding period.

\paragraph{2(d) Explicit license} MIT License

\paragraph{2(e) Structured metadata} \url{https://github.com/ebmlab/EvidenceOutcomes/blob/main/EvidenceOutcomes_meta_data_mlc.json}

\paragraph{2(f) Persistent dereferenceable identifier} \url{http://doi.org/10.5281/zenodo.10523337}